\RequirePackage{fix-cm}
\documentclass[smallcondensed]{svjour3}     
\smartqed  
\usepackage{graphicx}
\usepackage{natbib} 
\usepackage{amsfonts}
\usepackage{mathtools}
\usepackage{paralist}
\usepackage{hyphenat}
\usepackage{multirow}
\usepackage{tabularx}
\usepackage{booktabs}
\usepackage{color}
\usepackage{colortbl}
\usepackage{rotating}
\usepackage{xspace}
\usepackage{capt-of}

\makeatletter \let\cl@chapter\relax \makeatother 
\usepackage{cleveref}

\definecolor{LightCyan}{rgb}{0.88,1,1}
\definecolor{Gray}{gray}{0.9}

\newcommand{\NP}{\textsc{np}}

\newcommand{\gurobi}{\textsc{Gurobi}~$6.5$\xspace}

\newcounter{equationset}
\newcommand{\equationset}[1]{
  \refstepcounter{equationset}
  \noindent\makebox[\linewidth]{Model~\theequationset: #1}}

\newcounter{save}

\newenvironment{constraints}[1]
{\begin{enumerate}[(1)]\setcounter{enumi}{#1}}
{\end{enumerate}}

\newcommand{\model}[2]
{\setcounter{save}{\value{equation}}%
\begin{align}%
#1
\end{align}
\begin{constraints}{\value{save}}
#2
\end{constraints}}

\newcommand{\ctab}[5][table]{
  \begin{#1}[htb]
    \centering
    \caption{#2}
    \label{tab:#3}
    \begin{tabularx}{\textwidth}{#4}
      \toprule
      #5
      \bottomrule
    \end{tabularx}
  \end{#1}
}

\newcolumntype{C}{>{\centering\arraybackslash}X}%

\newcommand{\named}[6][\min]%
    { \begin{subequations}%
        \label{mod:#2}%
        \begin{align}%
        #1   && #4 \span \span \tag{#3} \\
        s.t. && #5 \\
             && #6 \span \span \nonumber
        \end{align}%
    \end{subequations}}

\journalname{EURO Journal on Transportation and Logistics}
\begin{document}

\title{Fleet Size and Mix Split-Delivery Vehicle Routing
}
\subtitle{A study of MIP formulations with CP integration}


\author{Arthur Mah\'eo
\and Tommaso Urli
\and Philp~Kilby
}

\authorrunning{A. Mah\'eo, T. Urli, and P. Kilby} 

\institute{The Australian National University \and
Data61, Canberra City \\
Level 3 Tower A, 7 London Circuit, City ACT 2601 \\
\email{\{Arthur.Maheo;Tommaso.Urli;Philip.Kilby\}@data61.csiro.au}
}

\date{Received: date / Accepted: date}

\hyphenation{Tru-deau}

\maketitle

\begin{abstract}
In the classic Vehicle Routing Problem (VRP) a fleet of of vehicles has to visit
a set of customers while minimising the operations' costs. We study a rich
variant of the VRP featuring split deliveries, an heterogeneous fleet, and
vehicle-commodity incompatibility constraints. Our goal is twofold: define the
cheapest routing and the most adequate fleet.

To do so, we split the problem into two interdependent components: a fleet
design component and a routing component. First, we define two Mixed Integer
Programming (MIP) formulations for each component. Then we discuss several
improvements in the form of valid cuts and symmetry breaking constraints.

The main contribution of this paper is a comparison of the four resulting models
for this Rich VRP. We highlight their strengths and weaknesses with extensive
experiments.

Finally, we explore a lightweight integration with Constraint Programming (CP).
We use a fast CP model which gives good solutions and use the solution to
warm-start our models.

\end{abstract}

\section{Introduction}

Beyond routing their vehicles at the least cost, goods carriers often want to
know what is the best fleet to use. For example, some vehicle types are more
expensive, in terms of maintenance and usage than others. Our work is based on a
tender for grocery delivery in Queensland, Australia. A goods carrier, handling
delivery of chilled and ambient temperature products, provided us with a year of
demand data. We want to determine what is the best fleet and the cheapest
routing to satisfy customers.

Our main contribution is to compare different modelling approaches, and provide
extensive computational experiments based on real-world instances. The ultimate
goal of this project is to find good modelling techniques for a more general
multi-day setting. In this work we focus on the single day setting as a
first step.

The Vehicle Routing Problem \citep[VRP,][]{Dantzig59,clarke1964scheduling}
consists in routing a fleet of vehicles to visit a set of customers. It was
first defined as an extension of the Travelling Salesman Problem (TSP) with
multiple salespeople. As such, it belongs to the class of \NP-hard problems
\citep{Lenstra81}. Its most common extension is the Capacitated VRP
\citep[CVRP,][]{Toth01}. In this case, customers have an associated demand and
vehicles have a limited capacity.

Another common extension is the Split Delivery VRP \citep[SDVRP,][]{Dror89}. In
the classic VRP, customers are visited exactly once. The SDVRP relaxes this
constraint, allowing more than one visit per customer. It is especially useful
when demands exceed the vehicle capacity. Although it is a relaxation, it has
been proven to be harder because because of the additional degree of freedom in
planning the deliveries. Despite its complexity, there are good reasons for
studying the SDVRP as it can lead to significant savings.

In recent years, driven by industrial needs, several realistic variants of the
VRP appeared. First, models were extended to handle multiple commodities (types
of goods). Usually such extension comes with compatibility issues, as vehicles
cannot transport all commodities. A more ambitious extension is the Fleet Size
and Mix VRP (FSMVRP) that considers finding a good fleet size and composition.

We call Rich VRP a problem merging many such extensions, see \citet{Caceres14}
for a survey. In this paper we consider the fleet size and mix, multi-commodity,
capacitated, and split delivery VRP.

Our current work tries to address issues raised in the experimental analysis of
\citet{Kilby16}. We propose a stronger MIP scheme to compare with their CP
approach.

To solve this problem, we focus on the two distinct, but interdependent parts of
the problem. First is the fleet design component, and second is the routing
component. Bigger vehicles with better compatibility are more expensive to
run. We propose two different MIP models for each component, amounting to four
complete MIPs.

Finally, we propose a lightweight integration of a CP model into our solving
process. The main strength of the CP model is to be able to find good solution
fast. Thus, we use it to prime our MIP search. Instead of letting the solver
find a starting solution, we use the CP model to find one.

\section{Related Work}
\label{sec:literature}

In this section we review some of the relevant literature on SDVRP, FSMVRP, and
RVRP formulations, and we consider some works on valid inequalities, some of
which we adapt to work with our models.

In \citet{Dror89} the authors present SDVRP together with an analysis of the
problem's complexity and the potential savings that can be attained by using it
as a delivery scheme. The conclusion of this work is that, despite the SDVRP
being a relaxation of the VRP, it is harder to solve. However, the additional
complexity pays off, since allowing split deliveries can lead to substantial
gains in terms of distances and number of vehicles needed, especially when the
demand of the customers is slightly less than a full truckload. Our work extends
the above, in that we consider additional compatibility constraints and a
multi-commodity scenario.

Regarding RVRP formulations we refer the interested reader to a recent,
excellent survey by \citet{Caceres14}.

Concerning exact algorithms to solve the VRP under capacity or time window
constraints, the interested reader can refer to \citet{Baldacci12}. The authors
present a comparison of different exact formulations. Two main class of
algorithms emerge: branch-and-cut algorithms and set covering algorithms. The
former are based on the idea of adding valid cuts during a branch-and-bound
search. The latter employ an exponential number of variables, and are therefore
solved using column generation. Our work can be seen as a preliminary study to
compare MIP models to be used in a decomposition scheme.

Regarding the research on valid inequalities for VRP, the literature is quite
rich. In an extension \citep{Dror94} of \citet{Dror89}, the authors focus on
improving the solution process by devising valid cuts for their previous model.
Both works are based on the traditional CVRP, where a single commodity is
considered, and the fleet is fixed.

In \citet{Letchford06}, a study on the relative strengths of different classes
of inequalities is presented, this time for the CVRP. In a follow-up to this
work \citep{Letchford15} the authors add two new multi-commodity formulations,
and study additional families of inequalities. In \citet{Baldacci12} a number of
valid cuts are also discussed; often these are not trivial as few of them have
polynomial algorithms available. Our approach adopts several of the valid
inequalities presented in the above works.

The FSMVRP is a variant of the Heterogeneous VRP (HVRP) where the number of
vehicles is not fixed. The HVRP has been first defined in the work of
\citet{Golden1984} as a relaxation of the CVRP where the vehicles do not have a
fixed type. The authors discuss a number of heuristic algorithms, some based on
the Clarke and Wright Savings algorithm, some based on the concept of
\emph{opportunity cost}, a way to evaluate the potential gain by using a vehicle
of a different type.

Most of the works on FSMVRP are limited to single-commodity and single-day
scenarios, furthermore very few exact algorithms exist in the literature, the
main focus being on improving the quality of linear relaxations in order to
strengthen bounding procedures -- e.g. \citet{Jabali2012}.


\citet{Vidal14} use a component-based framework to solve a variety real-world
VRP variants, including fleet size and mix vehicle routing, is proposed. The
approach, however, does not handle split deliveries or multiple commodities.

An exact algorithm was proposed by \citet{Baldacci2009} for the HVRP. Their
approach is based on a set partitioning formulation, which they solve using a
reduced model in combination with a bounding procedure. Although this approach
encompasses a wide variety of possible variations of the HVRP, it does not
handle split deliveries or multiple commodities.

\citet{Yaman2006} offers a work similar to ours in the sense that they design
and compare six different formulations for the HVRP: four based on the
Miller-Tucker-Zemlin constraints and two based on a flow formulation. They
design a number of valid inequalities for the different models, then report
computational experiments using standard benchmarks from the literature.
However, they only provide the time and value of the linear relaxations, no
exact result is provided.

For a recent review on different FSMVRP algorithms and problem classification,
we refer the interested read to a survey by \citet{kocc2016thirty}. Following
the classification proposed in another review by \citet{Baldacci2008a} our
problem is a FSMFD, with unlimited fleet size, fixed costs, and routing costs.

To the best of our knowledge the work of \citet{Kilby16} is the only combining
FSMVRP with split deliveries and multiple commodities. The authors present a CP
and Large Neighbourhood Search (LNS) approach to solve the same problem we
address here. The experimental analysis revealed the strength of MIP on some
particular problem configurations, and therefore motivated the present
research. Moreover, we employ the CP model presented in that work to prime our
MIP formulation, so as to increase the performance of our approach.

\section{Problem}
\label{sec:problem}
%
%
\newcommand{\dep}{\mathbf{dep}}%
\newcommand{\dem}{\mathbf{dem}}%
\newcommand{\cost}{\mathbf{cost}}%
\newcommand{\dist}{\mathbf{dist}}%
\newcommand{\capa}{\mathbf{cap}}%
\newcommand{\comp}{\mathbf{comp}}%
\newcommand{\comms}{\mathbf{K}}%
\newcommand{\comm}{k}%
\newcommand{\vtypes}{\mathbf{T}}%
\newcommand{\vtype}{t}%
\newcommand{\custs}{\mathbf{C}}%
\newcommand{\cust}{i}%
\newcommand{\custi}{\mathbf{C^\prime}}%
\newcommand{\edges}{\mathbf{E}}%
\newcommand{\edge}{e}%
\newcommand{\veh}{v}%
\newcommand{\vehs}{\mathbf{V}}%
\newcommand{\use}{u}%
In this section we provide a description of the problem components: entities,
constraints, and objective. We defer the description of routing and fleet
constraints to \Cref{sec:approach} as they are specific to each model. We
decided to limit the use of equations in this section because they are often
very similar. \Cref{ap:models} presents the complete mathematical formulation of
the resulting models which combine: the objective function, the core
constraints, one fleet model, and one routing model. \Cref{sec:complete}
presents an example augmented with the constraints.

\subsection{Entities}

\begin{table}[htb]
  \begin{tabularx}{\textwidth}{r X}
    $\dep$             & the depot where the vehicles start and end their routes; \\
    $\comms$           & the set of product types (or \emph{commodities}) that %
                         must be dispatched;  \\
    $\vtypes$          & the set of vehicle types that can be used to deliver the %
                         demand, which differ in capacity and cost;  \\
                         $\capa_\vtype$     & the total capacity of a type of vehicle (across all the commodities); \\
    $\custs$           & the set of customers to be served. We also denote by %
                         $\custi = \custs \cup \{\dep\}$ the set of all locations %
                         in the problem;  \\
    $\dem_{i,\comm}$   & the demand of commodity $\comm \in \comms$ by customer %
                         $i \in \custs$, we also denote by $\dem_i$ the total %
                         demand of customer $i$ (i.e. %
                         $\dem_i = \sum_{\comm \in \comms} \dem_{i, \comm}$);  \\
    $\comp_{t,\comm}$  & the compatibility between vehicle type %
                         $\vtype \in \vtypes$ and commodity $\comm \in \comms$
    \begin{equation*}
      \comp_{t,\comm} = \begin{cases}
        1 & \text{if vehicle type}~\vtype~\text{can transport commodity}~\comm,\\
        0 & \text{otherwise;}
      \end{cases}
    \end{equation*}  \\
    $\edges$           & the connections $\edge_{i,j},\;\forall i, j \in \custi$ %
                         between the locations; and \\
    $\cost_{t,i,j}$    & the cost of travelling on a connection $\edge_{i,j}$ with %
                         a vehicle of type $t$. \\
    $\dist_{i,j}$      & the the length of edge $\edge_{i, j}$. \\
  \end{tabularx}
  \label{tab:entities}
\end{table}

We consider the distribution network to be a graph $\mathbf{G} = \langle \custi,
\edges\rangle$ where the locations $\custi$ represent nodes and the connections
$\edges$ represent the edges. Moreover, all the models will refer to a set of
vehicles $\veh \in \vehs$ that can be either fixed or to be decided. We list the
different entities used to model the problem in \Cref{tab:entities}.


\subsection{Objective}

The problem we address is the daily component of an annual fleet size-and-mix
problem. The long term costs of running vehicles can be amortised to a daily
fixed cost, but we decided not to take them into account as:
\begin{inparaenum}[\itshape a)]
\item our partenr was not able to provide those costs;
\item they make little sense in the daily setting, although they are an easy
  extension.
\end{inparaenum}

Therefore the objective in our models is only to minimise the routing costs.

\subsection{Constraints}

We distinguish four types of \emph{core} constraints that will be present in
every problem variant.
\begin{description}
\item[\textbf{Demand satisfaction}]
  the demands of all customers must be satisfied, that is the sum of all
  deliveries of a given commodity to a customer must equal its demand.
\item[\textbf{Capacity constraints}]
  the capacities of vehicles must be respected, that is the sum of all
  deliveries by a single vehicle must not exceed its capacity.
\item[\textbf{Used vehicles}]
  unused vehicles stay at the depot, and therefore cannot travel on any
  edge or serve any customers.
\item[\textbf{Visited customers}]
  vehicles can only deliver customers that they visited.
\end{description}

\section{Our Approach}
\label{sec:approach}

In this section we present our approach, whose modularity emerges from the
natural decomposition of the addressed problem into fleet design and routing
components. In particular, we describe two models for each component, which can
be combined to assemble four different complete models. Moreover, we present a
number of valid cuts and symmetry breaking constraints that can be applied to
the models.

\subsection{Common Variables}

Our models share three sets of variables. The routing decisions are modelled
using the Boolean variables $x$ indicating whether a vehicle travels on a given
edge in the solution. The amount of each commodity delivered by a vehicle at a
customer is modelled by the continuous variables $y$. Finally, the Boolean
variables $\use$ represent the \emph{unused} vehicles, i.e., if $\use$ is equal
to one, it means that the vehicle is not used.

The usage variable comes from the yearly problem and tells whether a vehicle is
used on a given day. We decided to keep it in the daily setting as we found it
lead to stronger relaxations when deciding the fleet composition, it is also
used to formulate some valid cuts (e.g., \eqref{eq:max_vehs}).

\subsection{Fleet Models}

In this section we describe two different models to deal with the fleet design
component of the problem. In the \emph{Flexible} fleet, we have a fixed number
of trucks and use a Boolean variable $z$ to determine the type of each
vehicle. We can easily compute the maximum size of the fleet by considering only
the smallest trucks, then the question of the fleet composition remains. In the
\emph{Stable} fleet, we assign each vehicle a type beforehand and size the fleet
according to their respective characteristics.

\subsubsection{Flexible}

In addition to the core variables this model uses an additional set of Boolean
variables $z_{v,t}$ to indicate if vehicle $\veh \in \vehs$ is of type $\vtype
\in \vtypes$. In order to keep a linear model, as different vehicle types have
different costs, we have to decompose the routing decision on the type of the
vehicle and therefore use variables $x_{\veh, \vtype, i, j}$ to indicate if a
vehicle $\veh \in \vehs$ of type $\vtype \in \vtypes$ travels on edge
$\edge_{i,j} \in \edges$. As such, vehicles have access to networks of different
costs, but can only use the network of their chosen type. Finally, we need to
enforce compatibility because the type of each vehicle is not known in advance.
\model{%
    \sum_{i \in \custs} y_{v, i, k} & \leq \sum_{t \in \vtypes} z_{v, t} \cdot \capa_t \cdot \comp_{t, k} & \forall \veh \in \vehs, k \in \comms \\
    \sum_{t \in \vtypes} z_{v, t} &= 1 & \forall \veh \in \vehs \\
    \sum_{\edge_{i, j} \in \edges} x_{v, t, i, j} & \leq |\edges| \cdot z_{v, t} & \forall \veh \in \vehs, t \in \vtypes
}{%
    \item Compatibility constraint.
    \item All vehicles must have exactly one type.
    \item Vehicles can only use edges associated with their chosen type.
}

\subsubsection{Stable}

Instead of determining the composition of the fleet during the optimisation
another approach is to have vehicles with fixed types from the start. We obtain
a model with \emph{pools} of vehicles of a given type. We then need to size each
of these pools so that we are able to satisfy the demands.

Presetting the type of the vehicles allows us to use a three-index routing
variable, $x_{v, i, j}$, thus reducing greatly the number of variables in the
model.

Furthermore, we reduce the number of $y$ variables as compatibility becomes an
implicit characteristic. As it also eliminates infeasible deliveries, the
compatibility constraints can be removed. In the following we denote $\comms^v$
the set of compatible commodities for vehicle $v$. Finally, the $y$ variables
are bounded by the size of the vehicle instead of the maximum size in the fleet.

\subsection{Routing Models}

In this section we present two different models to deal with the routing aspects
of the problem. Both are based on classic formulations in the literature. The
first model is called \emph{Vehicle Flow}, reflecting the fact that the main
constraint models the balance in the in-flow and out-flow from a node.  The
second model is called \emph{Commodity Flow}, since it employs an additional
variable tracking the amount of goods carried on the edges.

\subsubsection{Vehicle Flow}

This model is very close to the standard 2-index model from the literature
\citet{Laporte83}, and requires a rather small number of variables. The
decision variables are the amount of flow on each vehicle between each pair of
customers.
\begin{align}
  \sum_{j \in \custi} x_{\veh,i,j} &= \sum_{j \in \custi} x_{\veh,j,i} & \forall \veh \in \vehs, i \in \custi
\end{align}

This formulation does not automatically eliminate sub-tours, and hence sub-tour
elimination constraints must be included. Because these are exponential in the
number of nodes, we use a classic sub-tour elimination procedure, which fires
whenever an integer solution is found during the branch-and-bound search. If we
find any sub-tour $S$ not including the depot, the following constraint is
added, with $\edges^S$ being the set of edges in S:
\begin{align}
  \sum_{\edge_{i,j} \in \edges^S} x_{\veh,i,j} & \leq |S|-1 & \forall S \subseteq \custs, \veh \in \vehs \label{eq:subtours}
\end{align}

\subsubsection{Commodity Flow}

This model uses a set of variables $f_{v, k, i, j}$ representing the amount of
commodity $\comm$ transported by vehicle $\veh$ on $\edge_{i,j}$
\citep[see][]{Gavish78}. As a consequence, the Commodity flow formulation has more
variables than the Vehicle flow formulation but results in a polynomial-sized model
since no sub-tour elimination constraints are required.
In order to guarantee flow consistency, we need to add the following set of
constraints:
\model{%
    \sum_{j \in \custi} \left( f_{\veh,\comm, j, i} - f_{\veh, \comm, i, j} \right) & = y_{\veh, i, \comm} & \forall \veh \in \vehs, i \in \custs, \comm \in \comms \\
    \sum_{\comm \in \comms} f_{\veh, \comm, i, j} & \leq x_{\veh, i, j} \cdot \capa_\veh & \forall \veh \in \vehs, \edge_{i,j} \in \edges
}{%
    \item The amount deposited at $i$ by vehicle $v$ is the difference between
      the amount carried before the visit and after the visit.
    \item A vehicle can only carry a load on the edges it uses.
}

\subsection{Valid Cuts}

\emph{Valid cuts} (or \emph{inequalities}) are an extremely effective technique
to solve MIP problems efficiently. Similarly to \emph{redundant constraints} in
CP, they are used to improve the strength of currently existing constraints
without removing optimal solutions. In this section we will present a set of
valid cuts derived from the literature and adapted to work in our rich VRP
context.

We do not present all variants of the constraints, only their most generic
expression. If a constraint differs considerably between two models, we present
the two formulations; otherwise, assume no changes apart from the indices.

\subsubsection{Minimum Visits}

This cut \citep{Dror94} provides a lower bound on the number of vehicles
required to visit a given customer, it is obtained by dividing its total demand
by the maximum capacity in our fleet.
\begin{align}
  \sum_{\veh \in \vehs} \sum_{j \in \custi} x_{\veh,j,i} & \geq \left\lceil\frac{\dem_{i}}{\capa_{max}}\right\rceil & \forall i \in \custs
\end{align}

\subsubsection{Minimum Vehicles}

This constraint, together with the branching strategy, was the most successful
improvement in the search for feasible solutions with a flexible fleet in
\citet{Kilby16}.

In the Vehicle model a similar cut can be implemented by ensuring that we will
have enough vehicle capacity to serve all demands.
\begin{align}
  \sum_{\veh \in \vehs} \sum_{\vtype \in \vtypes} z_{\veh, \vtype} \cdot \capa_\vtype & \geq \sum_{i \in \custs} \dem_{i,\comm} & \forall \comm \in \comms \label{eq:min_veh}
\end{align}

In the Stable model, since capacities are already fixed, we simply ensure that
enough vehicles leave the depot by using the $\use$ (usage) variables. This
provides an earlier cut in the branch-and-bound search since $\use$s have high
branching priority. Let $F_\vtype$ be the total number of vehicle of type
$\vtype$.
\begin{align}
  \sum_{\veh \in \vehs} \use_\veh & \leq \sum_{\vtype \in \vtypes} F_\vtype - \sum_{i \in \custs} \left\lceil\frac{\dem_{i}}{\capa_{max}}\right\rceil \tag{\ref{eq:min_veh}$^\prime$}
\end{align}

\subsubsection{Maximum Vehicles}

Conversely, when using the Stable fleet model, we can limit the number of
vehicle in operation by imposing a lower bound on the number $V^{\dep}$ of
vehicles staying at the depot.
\begin{align}
  \sum_{\veh \in \vehs} \use_\veh & \geq V^{\dep} \label{eq:max_vehs}
\end{align}

\subsubsection{Fractional Sub-Tours}

This cut \citep{Dror94} requires that if a vehicle visits a customer it has to
travel somewhere else afterwards.
\begin{align}
  x_{\veh, i, j} & \leq \sum_{\mathclap{l \in \custi, l \neq i}} x_{\veh, j, l} & %
  \forall \veh \in \vehs, \edge_{i, j} \in \edges
\end{align}

\subsubsection{Depot Outgoing Degree}

In its general form, this cut \citep{Dror94} states that all vehicles have to
exit the depot, however we amend it to allow unused vehicles to stay at the
depot.
\begin{align}
  \sum_{j \in \custs} x_{\veh, \dep, j} + \use_\veh & = 1 & \forall \veh \in \vehs
\end{align}

\subsubsection{Single visit.}
This cut enforces the fact that a vehicle can visit a customer at most once,
using the following symmetric constraints:
\begin{align}
  \sum_{j \in \custi} x_{\veh,i,j} & \leq 1 & \forall \veh \in \vehs, i \in \custs \\
  \sum_{j \in \custi} x_{\veh,j,i} & \leq 1 & \forall \veh \in \vehs, i \in \custs
\end{align}

\subsection{Symmetry Breaking Constraints}

\emph{Symmetry breaking} constraints try to alleviate the degeneracies of a
model by removing equivalent solutions. In this section we assume that the input
data, such as vehicles or customers, is ordered as integer sequences. In
particular, the depot would be customer $0$.

\subsubsection{Usage}

If a vehicle is unused, the \emph{next} vehicle must be unused too so as to
avoid recombination to determine the number of used vehicles.
\begin{align}
  \use_{\veh-1} & \geq \use_{\veh} & \forall \veh \in \vehs
\end{align}

\subsubsection{Visit Order}

We impose that a vehicle $\veh$ can visit customer $i$ only if vehicle $\veh-1$
has visited any location in $[1..i]$.

\begin{align}
  \sum_{h \in \custi} \sum_{l = 1}^j x_{\veh - 1, h, l} & \geq x_{\veh, i, j} & \forall \veh \in \vehs, i \in \custs, j \in \custi \label{eq:visit_order}
\end{align}

In the Flexible model this must be restricted to vehicles of the same type.
\begin{align}
  \sum_{u = 0}^t z_{\veh-1, u} + \sum_{h \in \custs} \sum_{l = 1}^{j} x_{\veh - 1, \vtype, h, l} & \geq x_{\veh, \vtype, i, j} & \forall \veh \in \vehs, \vtype \in \vtypes, i \in \custs, j \in \custi \tag{\ref{eq:visit_order}$'$}
\end{align}

\subsubsection{Fleet Order}

When using the Flexible model, we force the vehicle types to be ordered, that is
if vehicle $\veh$ is of type $\vtype$, then vehicle $\veh + 1$ can only take
types $[t .. T]$.
\begin{align}
  \sum_{\tau = 0}^{\vtype} z_{\veh - 1, \tau} & \geq z_{\veh, \vtype} & \forall \veh \in \vehs, \vtype \in \vtypes
\end{align}

\subsubsection{Customer Assignment}

Inspired by \citet{Dror94}, where the authors design a cut assigning the first
vehicle to the farthest customer, we re-order the input data so that the farther
customers (from the depot) are assigned smaller indices.

Using a Flexible fleet, with $\hat{\jmath}$ being the farthest customer, we
have:
\begin{align}
  \sum_{i \in \custs} \sum_{t \in \vtypes} x_{0, t, i, \hat{\jmath}} = 1 \label{eq:farthest_customer}
\end{align}

With a Stable fleet we can only have a weaker constraint where we force at least
one of the first vehicles of each type to visit the farthest customer. With
$\Pi$ being the sets of such vehicles.
\begin{align}
  \sum_{\veh \in \Pi} \sum_{i \in \custs} x_{\veh, i, \hat{\jmath}} & \geq 1 \tag{\ref{eq:farthest_customer}$^\prime$}
\end{align}

\subsubsection{Total Load}

In the Commodity flow model we also enforce that all vehicles have to leave the
depot fully loaded. However, we do not constrain their final load as this can be
adjusted easily.
\begin{align}
\sum_{\comm \in \comms} \sum_{j \in \cust} f_{\veh, \comm, \dep, j} & = \capa_\veh & \forall \veh \in \vehs
\end{align}

\subsection{Complete Example}
\label{sec:complete}

We present here a complete mathematical formulation using: Commodity routing
model (\Cref{eq:balance,eq:carry}) and Stable fleet model (e.g. implicit
compatibility with $\comms^\veh$). The model is composed of: the objective
function, core constraints (\Cref{eq:demand,eq:capacity,eq:usage,eq:visited}),
valid cuts (\Cref{eq:nb_visits,eq:min_vehs,eq:max_vehs,eq:frac_subtour,%
  eq:depot_exit,eq:depot_load,eq:single_entry,eq:single_exit}), symmetry
breaking constraints (\Cref{eq:use_order,eq:cust_order}), and ordering
constraints (\Cref{eq:farthest_visit}).

\named{mip}%
      {MIP}%
      {\sum_{\veh \in \vehs} \sum_{\edge_{i, j} \in \edges} x_{\veh, i, j} \cdot \cost_\veh \cdot \dist_{i, j} \label{eq:objective}}%
      {\sum_{\veh \in \vehs} y_{\veh, \cust, \comm} & = \dem_{\cust, \comm} & \forall \cust \in \custs, \comm \in \comms \label{eq:demand} \\
        && \sum_{\cust \in \custs} \sum_{\comm \in \comms^\veh} y_{\veh, \cust, \comm} & \leq \capa_{\veh} & \forall \veh \in \vehs  \label{eq:capacity} \\
        && \sum_{\edge_{i,j} \in \edges} x_{\veh, i, j} & \leq u_{\veh} & \forall \veh \in \vehs  \label{eq:usage} \\
        && \sum_{j \in \custs} x_{\veh, j, \cust} \cdot \dem_{\cust, \comm} & \geq y_{\veh, \cust, \comm} & \forall \veh \in \vehs, i \in \custs, \comm \in \comms^\veh \label{eq:visited} \\
        && \sum_{j \in \custi} (f_{\veh,\comm, j, i} - f_{\veh, \comm, i, j}) & =  y_{\veh, \cust, \comm}  & \forall \veh \in \vehs, i \in \custs,  \comm \in \comms ^\veh \label{eq:balance} \\
        && \sum_{\comm \in \comms^\veh} f_{\veh,\comm, i, j} & \leq x_{\veh, i, j} \cdot \capa_\veh & \forall \veh \in \vehs, \edge_{i,j} \in \edges  \label{eq:carry} \\
        && \sum_{\veh \in \vehs} \sum_{j \in \custi} x_{\veh,j,i} & \geq \left\lceil \frac{\dem_{i}}{\capa_{max}} \right\rceil & \forall i \in \custs  \label{eq:nb_visits} \\
        && \sum_{\veh \in \vehs} \use_\veh & \leq \sum_{\vtype \in \vtypes} F_\vtype - \sum_{i \in \custs} \left\lceil \frac{\dem_{i}}{\capa_{max}} \right\rceil \span  \label{eq:min_vehs} \\
        && \sum_{\veh \in \vehs} \use_\veh & \geq V^{\dep}  \label{eq:max_vehs} \\
        && x_{\veh, i, j} & \leq \sum_{\mathclap{l \in \custi, l \neq i}} x_{\veh, j, l} & \forall \veh \in \vehs, \edge_{i, j} \in \edges  \label{eq:frac_subtour} \\
        && \sum_{j \in \custs} x_{\veh, \dep, j} + \use_\veh & = 1 & \forall \veh \in \vehs  \label{eq:depot_exit} \\
        && \sum_{j \in \custs} \sum_{\comm \in \comms^\veh} f_{\veh, \comm, \dep, j} &  = \capa_\veh & \forall \veh \in \vehs  \label{eq:depot_load} \\
        && \sum_{j \in \custi} x_{\veh,i,j} & \leq 1 & \forall \veh \in \vehs, i \in \custs  \label{eq:single_entry} \\
        && \sum_{j \in \custi} x_{\veh,j,i} & \leq 1 & \forall \veh \in \vehs, i \in \custs  \label{eq:single_exit} \\
        && \use_{\veh-1} & \geq \use_{\veh} & \forall \veh \in \vehs  \label{eq:use_order} \\
        && \sum_{h \in \custi} \sum_{l = 1}^j x_{\veh - 1, h, l} & \geq x_{\veh, i, j} & \forall \veh \in \vehs, i \in \custs, j \in \custi  \label{eq:cust_order} \\
        && \sum_{\veh \in \Pi} \sum_{i \in \custs} x_{\veh, i, \hat{\jmath}} & \geq 1  \label{eq:farthest_visit}
      }%
      {x \in \mathbb{B}, u \in \mathbb{B}, y \geq 0, f \geq 0}

\section{Experimental Comparison} \label{sec:expes}

In this section we compare several variants of the presented models. Each
experiment is run for $15$ minutes on an AMD Opteron $4334$ machine running at
$3.1$ GHz and with $64$ GB of RAM. The models were implemented in \gurobi
\citep{gurobi}, and run using the default parameters except a higher branching
priority for the usage variable for all models, and then for the truck type in
the Flexible variant.

We have three groups of instances: small, medium, and large. The latter
represent real demand data, whereas in medium and small a different degree of
clustering is used to aggregate the demands of nearby customers. On average,
there are $5$ customer nodes in a small instance, $15$ in a medium, and $18$ in
a large.

To size the fleet, we rely on a characteristic of our case study: refrigerated
vehicles are more expensive to run than regular vehicles. Hence, we only
consider chilled demand for determining the number of refrigerated vehicles
required, and ambient demand for regular vehicles.

In our results we use the following abbreviations: the first letter is for the
fleet model (\emph{s} for Stable and \emph{f} for Flexible); the second is for
the routing model (\emph{c} for Commodity and \emph{v} for Vehicle).

\subsection{MIP Comparison}
\label{sec:mip_results}

This first set of experiments reports the results obtained over nine small
instances by the four MIP models, augmented with valid cuts, symmetry
constraints, and ordering constraints. Each instance was run for a maximum of
fifteen minutes. We report the best objective value found across all variants;
then the run time in seconds (if the time limit was not exceeded) or the final
integrality gap (if the time limit was hit) for each of the variants; and
finally the number of sub-tour elimination constraints \eqref{eq:subtours} added
in the two models based on Vehicle flow. The last line recaps the average gap
found on instances not solved to optimality. We highlight rows where optimality
was not reached by any model as well as the best time or gap per row.

The tables in \Cref{ap:results} provide the results for the complete set of
experiments, starting with the standard MIP formulation as presented in
\citet{Kilby16}. \Cref{tab:order} presents the results of the most complete
model which consists of: valid cuts, symmetry breaking constraints, and
ordering.

In \Cref{tab:relaxation} we provide the time taken in seconds and the gap
between the optimal solution, or best known solution, and the linear relaxation;
and respectively with the first heuristic solution in \Cref{tab:heuristic}.

\ctab{Results of using different models. For each instance we report the best value found, time taken or integrality gap at $15$ min, and number of sub-tour elimination constraints. The best results are in bold. Instances where the optimum is not reached are greyed.}
     {order}
     {*{8}{C}}
     {
     \multirow{2}{*}{Instance} & \multirow{2}{*}{Value} & \multicolumn{4}{c}{Time (s) or gap at 900s} & \multicolumn{2}{c}{\# subtours \eqref{eq:subtours}} \\
     \cmidrule(lr){3-6} \cmidrule(lr){7-8}
     & & sc & sv & fc & fv & sv & fv \\
     \midrule
     1 & 31358.95 & \bfseries 15.19 & 26.02 & 32.63 & 33.81 & 216 & 250 \\
     2 & 45538.02 & \bfseries 723.32 & 758.51 & (0.20\%) & (1.13\%) & 196 & 325 \\
     3 & 54150.16 & \bfseries 70.89 & 343.70 & 503.72 & (0.62\%) & 456 & 1020 \\
     4 & 45435.52 & \bfseries 326.36 & 680.91 & 505.64 & (1.30\%) & 341 & 490 \\
     5 & 53989.22 & \bfseries 817.45 & (1.02\%) & (1.27\%) & (1.79\%) & 117 & 255 \\
     \rowcolor{Gray}
     6 & 60943.93 & \bfseries (0.11\%) & (1.55\%) & (0.83\%) & (3.13\%) & 765 & 1700 \\
     \rowcolor{Gray}
     7 & 62864.21 & \bfseries (0.62\%) & (1.81\%) & (1.12\%) & (3.77\%) & 430 & 1700 \\
     8 & 23219.89 & \bfseries 2.25 & 2.35 & 25.46 & 22.99 & 0 & 0 \\
     9 & 44738.91 & \bfseries 125.65 & 259.11 & 281.49 & 550.84 & 203 & 140 \\
     \cmidrule{2-6}
     \multicolumn{2}{r}{Avg. gap} & \bfseries 0.36\% & 1.46\% & 0.85\% & 1.96\% \\
     }

\ctab{Root node relaxation results. The Time columns give the time taken to solve the relaxed problem; the Gap columns give the percentage difference with the best known value. The smallest gaps are in bold. Instances where optimum is not reached are greyed.}
     {relaxation}
     {c *{8}{C}}
     {
       \multirow{2}{*}{Instance} & \multicolumn{2}{c}{sc} & %
       \multicolumn{2}{c}{sv} & \multicolumn{2}{c}{fc} & \multicolumn{2}{c}{fv} \\
       \cmidrule(lr){2-3} \cmidrule(lr){4-5} \cmidrule(lr){6-7} \cmidrule(lr){8-9}
       & Time (s) & Gap (\%) & Time (s) & Gap (\%) & Time (s) & Gap (\%) %
       & Time (s) & Gap (\%)  \\
       \midrule
       1 & 0.44 & \bfseries 6.61 & 0.35 & 88.64 & 2.04 & 8.53 & 1.18 & 86.36 \\
       2 & 0.52 & \bfseries 8.41 & 0.29 & 81.05 & 3.07 & 8.92 & 1.23 & 99.98 \\
       3 & 1.09 & \bfseries 3.35 & 0.48 & 112.97 & 4.99 & 4.13 & 3.02 & 136.74 \\
       4 & 0.73 & \bfseries 3.43 & 0.45 & 112.15 & 3.58 & 3.56 & 1.92 & 138.54 \\
       5 & 1.05 & 4.38 & 0.42 & 83.89 & 4.68 & \bfseries 4.12 & 2.16 & 128.39 \\
       \rowcolor{Gray}
       6 & 1.10 & \bfseries 4.47 & 0.61 & 111.09 & 6.86 & 4.57 & 3.48 & 131.70 \\
       \rowcolor{Gray}
       7 & 1.47 & \bfseries 4.55 & 0.50 & 111.39 & 6.11 & 4.88 & 4.12 & 153.74 \\
       8 & 0.10 & 13.99 & 0.07 & 14.36 & 0.67 & \bfseries 13.22 & 0.40 & 77.54 \\
       9 & 0.66 & \bfseries 8.05 & 0.33 & 79.06 & 2.73 & 8.86 & 1.59 & 95.60 \\
       \midrule
       \multicolumn{2}{l}{Avg. gap (\%)} & \bfseries 6.89 && 81.73 && 7.33 && 109.02 \\
     }

\ctab{First integer solution results. The Time columns give the time taken to find the first integer solution; the Gap columns the percentage difference with the best known value. The smallest gaps are in bold. Instances where optimum is not reached are greyed.}
     {heuristic}
     {*{9}{C}}
     {
       \multirow{2}{*}{Instance} & \multicolumn{2}{c}{sc} & %
       \multicolumn{2}{c}{sv} & \multicolumn{2}{c}{fc} & \multicolumn{2}{c}{fv} \\
       \cmidrule(lr){2-3} \cmidrule(lr){4-5} \cmidrule(lr){6-7} \cmidrule(lr){8-9}
       & Time (s) & Gap (\%) & Time (s) & Gap (\%) & Time (s) & Gap (\%) %
       & Time (s) & Gap (\%)  \\
       \midrule
       1 & 0.15 & 56.16 & 0.08 & 54.20 & 0.46 & 55.85 & 0.44 & \bfseries 42.76 \\
       2 & 0.10 & 54.63 & 0.01 & 50.24 & 9.27 & 13.84 & 8.99 & \bfseries 13.00 \\
       3 & 0.15 & 56.68 & 0.38 & 37.90 & 7.13 & \bfseries 34.25 & 2.53 & 38.10 \\
       4 & 0.10 & 55.43 & 0.03 & 45.64 & 16.71 & \bfseries 21.59 & 1.69 & 27.01 \\
       5 & 0.16 & 56.08 & 0.11 & 32.80 & 8.00 & 32.46 & 1.89 & \bfseries 31.66 \\
       \rowcolor{Gray}
       6 & 0.16 & 57.59 & 0.01 & 49.69 & 29.16 & \bfseries 10.33 & 2.99 & 32.88 \\
       \rowcolor{Gray}
       7 & 0.18 & 57.31 & 0.39 & 39.54 & 64.35 & \bfseries 19.13 & 3.62 & 38.09 \\
       8 & 0.01 & 44.76 & 0.00 & \bfseries 36.79 & 0.19 & 47.12 & 0.14 & 44.07 \\
       9 & 0.14 & 58.67 & 0.01 & 53.48 & 23.59 & \bfseries 17.76 & 1.24 & 29.88 \\
       \midrule
       \multicolumn{2}{l}{Avg. gap (\%)} & 54.63 && 44.44 && \bfseries 31.84 && 32.35 \\
     }

From these results we can clearly see that a model dominates the others: the
Stable fleet with Commodity routing, Model SC, presented in its augmented
version as Model \eqref{mod:mip}. First of all, predefining the types of each
vehicle instead of having to determine their type during the optimization is the
biggest improvement. While it increases the number of vehicles to route it
decreases the number of decision variables.
Secondly, having a polynomial-sized formulation by using a flow-based
formulation for the routing gives far better results than relying on sub-tour
elimination. This is explained by the fact that such a formulation has a better
linear relaxation (see \Cref{tab:relaxation}), and therefore a smaller search
tree. Conversely, the reason why the first integer solution for flow based
models is better is because they need to add sub-tour elimination constraints to
reach feasibility, and it is reflected in the time they take to find that first
solution.

This result concurs with those of \citet{Yaman2006} whose best performing model
is based on a flow model and fleet is always predefined.

\subsection{Warm-Starting the MIP Using CP}

In this second set of experiments we attempt to integrate CP to MIP. We devised
a rather lightweight integration between the CP solver in \citet{Kilby16} and
our MIP models. Since the CP-based solver returns a reasonably good solution
quickly, the idea is: first, find a solution with CP; then use that solution as
a starting solution for the MIP; and, from there, let the MIP search proceed
normally. We call this: \emph{warm starting} the MIP.

In the following experiments, we keep the same $15$-minutes time limit as
before, but we allow a variable amount of time (ranging between $0$ and $15$
minutes) to the CP solver to find an initial solution. In order to keep the
length of the experiments constant, the longer the CP search runs, the shorter
the time limit for the MIP search.

The CP solver functions by finding a first feasible solution and then proceeding
to improve it by means of an LNS procedure. Therefore, in our experiments,
allocating a time of $0$ seconds to the CP solver means we use the first
feasible solution it finds instead of letting the CP solver improve it.

\begin{figure}[ht]
\centering
\small
\caption{Evolution of final gap when using CP solution, based on a single
  instance, day 6, for the four models. The $x$ axis is the time allotted to the
  CP model, then the MIP has the remaining of the fifteen minutes. A dot means
  the MIP finished. A missing curve means that the LP relaxation at the root
  node did not finish under the remaining time.}
\resizebox{\columnwidth}{!}{\input{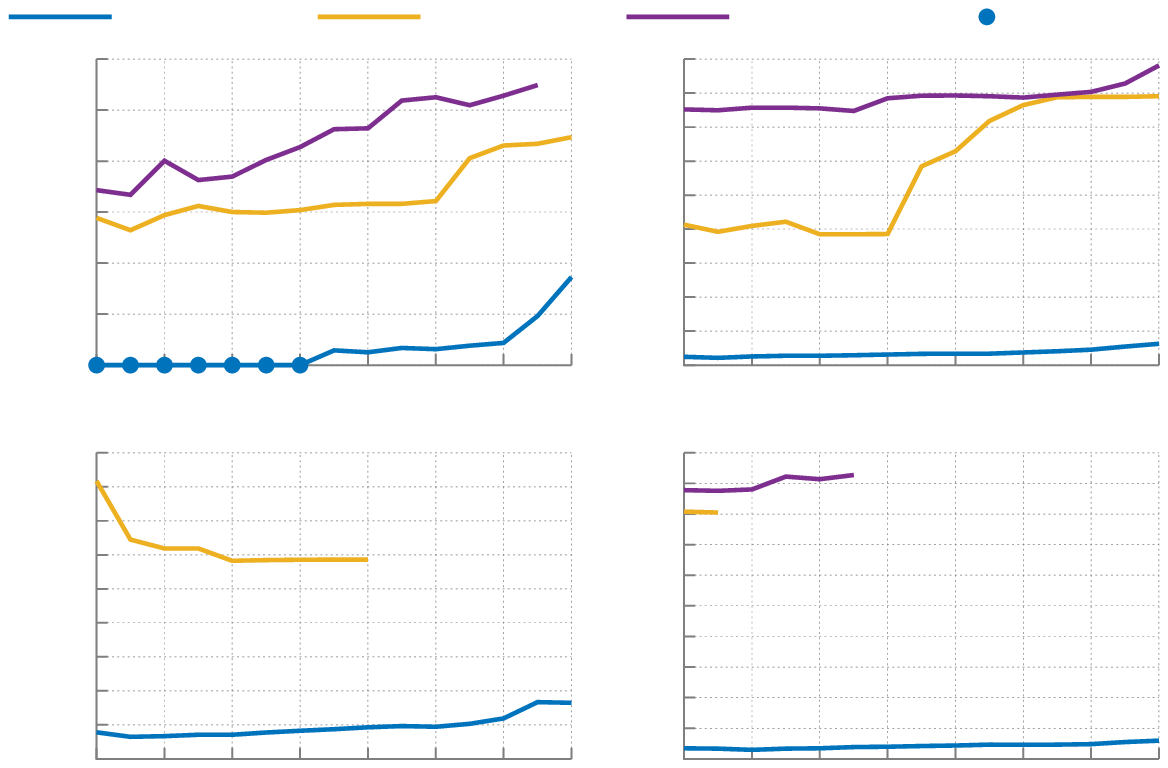}}
\label{fig:gap}
\end{figure}

\Cref{fig:gap} presents the evolution of the final MIP gap on a selected
instance, day 6, for all instance sizes and model variations. We report the
final integrality gap function of the time allocated to CP.

The slight dip early in \Cref{fig:gap} shows that the first solution found by CP
is not the most efficient for warm starting the MIP search, and letting CP
improve it for about one minute yields better results. Allowing longer CP
run-times decreases the overall solution quality. The reason behind this is that
CP finds a feasible solution quickly, then has to spend more and more time
improving it as the solutions get better. At a given point, letting MIP improve
the solution proves to be a better strategy than letting CP do it. For the
flexible fleet with commodity flow, it is even impossible to find even the
solution to the linear relaxation at the root node under fifteen minutes on
large instances.

\Cref{tab:prime} reports the results of warm-starting the MIP using a solution
found by the CP model in 1 minute; it presents the results in a similar fashion
to those in \Cref{sec:mip_results}. Do note that the time reported is for the
MIP only, not taking the CP runtime into account.

\ctab{Warm-starting MIP with a CP solution found in 1 minute}%
     {prime}%
     {*{8}{C}}
     {
     \multirow{2}{*}{Instance} & \multirow{2}{*}{Value} & \multicolumn{4}{c}{Time (s) or gap at 900s} & \multicolumn{2}{c}{\# subtours \eqref{eq:subtours}} \\
     \cmidrule(lr){3-6} \cmidrule(lr){7-8}
     & & sc & sf & fc & ff & sf & ff \\
     \midrule
     1 & 31358.95 & 29.31 & 27.82 & \bfseries 19.83 & 58.84 & 120 & 300 \\
     2 & 45538.02 & \bfseries 522.84 & 806.01 & 598.22 & (0.36\%) & 196 & 260 \\
     3 & 54150.16 & \bfseries 72.99 & 664.07 & 506.86 & (0.26\%) & 342 & 340 \\
     4 & 45435.52 & 222.91 & \bfseries 207.41 & 438.87 & 517.95 & 62 & 770 \\
     5 & 53989.22 & \bfseries 447.78 & (1.48\%) & (1.37\%) & (1.65\%) & 78 & 425 \\
     6 & 60943.93 & \bfseries 664.57 & (2.15\%) & (1.10\%) & (3.51\%) & 270 & 800 \\
     \rowcolor{Gray}
     7 & 62864.21 & \bfseries (0.45\%) & (1.09\%) & (1.53\%) & (3.16\%) & 301 & 1200 \\
     8 & 23219.89 & \bfseries 4.58 & 4.61 & 27.21 & 18.78 & 36 & 40 \\
     9 & 44738.91 & \bfseries 160.73 & 183.37 & 161.06 & 438.17 & 261 & 770 \\
     \cmidrule{2-6}
     \multicolumn{2}{r}{Avg. gap} & \bfseries 0.45\% & 1.58\% & 1.34\% & 1.79\% \\
     }

Warm starting the MIP search with a CP solution gives satisfying results: one
more instance is now solved to optimality under the 15 minutes constraint;
overall solving times are reduced; and, although the average gap reported on the
last line is higher for some models -- e.g. SC -- this comes from having less
unsolved days -- and then the gap is lower, $0.62\%$ vs. $0.45\%$.

On the other hand, we can observe a slight decrease in performance in some
cases, with higher time or final gaps. This comes from the \emph{starting
  location} for the MIP; in other words, imposing a starting solution to the MIP
model makes the search start at a given position in the tree, with no warranty
that the optimal solution is close. In that case, letting the MIP model do the
exploration and pruning may lead to a smaller search tree as the region of the
initial solution will never be explored.

We can observe that most model find the optimal solution early on but fail to
prove its optimality, this is shown by the best value reported for unsolved
instances being equal to the optimal value. Using the CP to warm start the
search helps reducing the size of the search tree, but not always. A way to
improve the solving process would be to determine which solutions are
interesting, or find a way to close the integrality gap faster.

\subsection{Results Over a Year}

In the final experiment from \citet[Table 2]{Kilby16} the authors compared the
results of two CP models, one with a standard B\&B search and one with the
custom LNS search, and their MIP model using one year of real world demand
provided by the client.

We ran the same experiment using the MIP model \eqref{mod:mip}. \Cref{tab:year}
summarizes the results on all instances, grouped by quarter, of: the CP model
using LNS search, our MIP, and our MIP warm-started with a CP solution obtained
in one minute.

For each quarter, we report the average value found across all the instances,
how often the model produced the best known solution, the percentage difference
with the best found value for each instance, and the final gap for the MIP and
the combination CP+MIP.

\ctab{Summary of results using one year of data, grouped by quarter, for each instance size. For each approach we have the average value over all instances, the percentage of instances where the approach found the best known value, and the percentage difference with the best known value. We also report the average integrality gap for the MIP based approaches.}
     {year}
    {p{0.5em} p{0.25em} c C C *{2}{c C C C}}
     {
       && \multicolumn{3}{c}{CP} & \multicolumn{4}{c}{MIP} & \multicolumn{4}{c}{CP (1min) + MIP} \\
       \cmidrule(lr){3-5} \cmidrule(lr){6-9} \cmidrule(lr){10-13}
       && \multirow{2}{*}{Value} & Best & Diff. %
       & \multirow{2}{*}{Value} & Best & Diff. & Gap %
       & \multirow{2}{*}{Value} & Best & Diff. & Gap  \\
       && & (\%) & (\%) && (\%) & (\%) & (\%) && (\%) & (\%) & (\%) \\
       \midrule
       \multirow{3}{*}{Q1} & S & 48959 & 0.00 & 0.78 & 48582 & 7.78 & \bfseries 0.00 & \bfseries 0.81 & \bfseries 48582 & \bfseries 92.22 & 0.02 & 0.82 \\
       & M & 51353 & 7.78 & 3.59 & 51320 & 40.00 & 3.50 & 7.39 & \bfseries 50068 & \bfseries 52.22 & \bfseries 0.98 & \bfseries 6.03 \\
       & L & 52406 & 6.67 & 4.96 & 52455 & 27.78 & 5.00 & 10.17 & \bfseries 50456 & \bfseries 65.56 & \bfseries 1.33 & \bfseries 7.46 \\
       \midrule
       \multirow{3}{*}{Q2} & S & 47511 & 1.11 & 1.06 & 47030 & 7.78 & 0.01 & \bfseries 0.68 & \bfseries 47026 & \bfseries 91.11 & \bfseries 0.01 & 0.76 \\
       & M & 49896 & 5.56 & 4.06 & 48954 & \bfseries 47.78 & 2.42 & 6.50 & \bfseries 48634 & 46.67 & \bfseries 1.59 & \bfseries 6.27 \\
       & L & 50956 & 6.67 & 5.45 & 50602 & 27.78 & 4.43 & 9.89 & \bfseries 48921 & \bfseries 65.56 & \bfseries 1.32 & \bfseries 7.45 \\
       \midrule
       \multirow{3}{*}{Q3} & S & 49043 & 0.00 & 1.04 & 48534 & 7.69 & \bfseries 0.00 & \bfseries 0.74 & \bfseries 48532 & \bfseries 92.31 & 0.01 & 0.80 \\
       & M & 51359 & 5.49 & 3.72 & 50551 & 37.36 & 2.37 & 6.88 & \bfseries 50062 & \bfseries 57.14 & \bfseries 1.61 & \bfseries 6.20 \\
       & L & 52532 & 8.79 & 5.19 & 52090 & 26.37 & 4.07 & 9.99 & \bfseries 50652 & \bfseries 64.84 & \bfseries 1.75 & \bfseries 8.02 \\
       \midrule
       \multirow{3}{*}{Q4} & S & 51700 & 2.20 & 0.93 & 51261 & 12.09 & \bfseries 0.00 & \bfseries 0.73 & \bfseries 51259 & \bfseries 85.71 & 0.01 & 0.77 \\
       & M & 54182 & 12.09 & 3.94 & 53972 & 30.77 & 2.88 & 7.12 & \bfseries 53022 & \bfseries 57.14 & \bfseries 1.65 & \bfseries 6.00 \\
       & L & 54179 & 8.79 & 4.72 & 53654 & 23.08 & 3.17 & 9.34 & \bfseries 52557 & \bfseries 67.03 & \bfseries 1.85 & \bfseries 7.81 \\
       \midrule
       \multirow{3}{*}{Y} & S & 49303 & 0.83 & 0.95 & 48852 & 8.83 & \bfseries 0.00 & \bfseries 0.74 & \bfseries 48850 & \bfseries 90.34 & 0.01 & 0.79 \\
       & M & 51698 & 7.73 & 3.83 & 51199 & 38.98 & 2.79 & 6.97 & \bfseries 50446 & \bfseries 53.29 & \bfseries 1.46 & \bfseries 6.12 \\
       & L & 52518 & 7.73 & 5.08 & 52200 & 26.25 & 4.17 & 9.85 & \bfseries 50646 & \bfseries 65.74 & \bfseries 1.56 & \bfseries 7.68 \\
     }

The combination CP+MIP consistently dominates its individual components. Even
though the MIP sometimes reaches a better value in the end, especially on medium
size instances, overall the combination CP+MIP has a lower value overall, as
shown by the lower percentage difference overall. The combined models also have
a lower integrality gap after 15 minutes. Furthermore, the combined CP+MIP has a
greater success rate in finding the optimal under the time limit ($64\%$)
compared to the simple MIP ($48\%$).

Instances \emph{better} solved using the MIP without warm-start are symptomatic
of those cases where the CP solution, albeit of good quality, is \emph{far} from
the actual optimal and the MIP spends a lot of time backtracking in the search
tree. In \emph{small} instances, the MIP has a lower percentage difference and
better gap because it solves less instances to optimality, therefore the
averages are lower.

\section{Conclusions}

We presented various MIP formulations for the fleet size and mix,
multi\hyp{}commodity, split-delivery vehicle routing problem with compatibility
constraints. Our formulation is modular, in that we split the fleet design and
the routing components of the problem and we develop two models for each
component.

We carried out an extensive experimental analysis to compare the MIP models, and
to measure their performance with respect to the CP-based LNS approach presented
in \citet{Kilby16}. Overall, the Stable fleet model and Commodity routing model
combination seem to attain the best performance, outperforming the LNS approach
on most instances, and represents the most promising candidate for a further
exploration of decomposition models for the multi-day extension of the problem.
Within our experimental analysis, we have also explored a lightweight
integration of CP and MIP, which consists in priming the MIP model with a
solution found by CP, replacing the heuristic solution found by the standard MIP
solver (in our case \gurobi). The effect on the integrality gap is positive,
however enough time must be allotted to the MIP search procedure to find a
better solution.


\bibliographystyle{spbasic}      

\newpage

\appendix

\section{Models}
\label{ap:models}

In this section we will present the different models created, without additional
valid cuts or symmetry breaking constraints.

\equationset{Stable Fleet and Commodity Flow}
\named{sc}%
      {SC}%
      {\sum_{\veh \in \vehs} \sum_{\edge_{i, j} \in \edges} x_{\veh, i, j} \cdot \mathbf{cost}_\veh \cdot \mathbf{dist}_{i, j}}%
      {\sum_{\veh \in \vehs} y_{\veh, \cust, \comm} & = \dem_{\cust, \comm} & \forall \cust \in \custs, \comm \in \comms \\
        && \sum_{\cust \in \custs} \sum_{\comm \in \comms^\veh} y_{\veh, \cust, \comm} & \leq \capa_{\veh} & \forall \veh \in \vehs \\
        && \sum_{\edge_{i,j} \in \edges} x_{\veh, i, j} & \leq u_{\veh} & \forall \veh \in \vehs \\
        && y_{\veh, \cust, \comm} & \leq \sum_{j \in \custs} x_{\veh, j, \cust} \cdot \dem_{\cust, \comm} & \forall \veh \in \vehs, i \in \custs, \comm \in \comms^\veh \\
        && \sum_{j \in \custi} (f_{\veh,\comm, j, i} - f_{\veh, \comm, i, j}) & = y_{\veh, i, \comm} & \forall \veh \in \vehs, i \in \custs, \comm \in \comms^\veh \\
        && \sum_{\comm \in \comms^\veh} f_{\veh,\comm, i, j} & \leq x_{\veh, i, j} \cdot \capa_\veh & \forall \veh \in \vehs, \edge_{i,j} \in \edges}%
      {x \in \mathbb{B}, u \in \mathbb{B}, y \geq 0, f \geq 0}

\equationset{Stable Fleet and Vehicle Flow}
\named{sf}%
      {SF}%
      {\sum_{\veh \in \vehs} \sum_{\edge_{i, j} \in \edges} x_{\veh, i, j} \cdot \mathbf{cost}_\veh \cdot \mathbf{dist}_{i, j}}%
      {\sum_{\veh \in \vehs} y_{\veh, \cust, \comm} & = \dem_{\cust, \comm} & \forall \cust \in \custs, \comm \in \comms \\
        && \sum_{\cust \in \custs} \sum_{\comm \in \comms^\veh} y_{\veh, \cust, \comm} & \leq \capa_{\veh} & \forall \veh \in \vehs \\
        && \sum_{\edge_{i,j} \in \edges} x_{\veh, i, j} & \leq u_{\veh} & \forall \veh \in \vehs \\
        && y_{\veh, \cust, \comm} & \leq \sum_{j \in \custs} x_{\veh, j, \cust} \cdot \dem_{\cust, \comm} & \forall \veh \in \vehs, i \in \custs, \comm \in \comms^\veh \\
        && \sum_{j \in \custi} x_{\veh, i, j} & = \sum_{j \in \custi} x_{\veh, j, i} & \forall \veh \in \vehs, i \in \custi \\
        && \sum_{\edge_{i, j} \in \edges^S} x_{\veh, i, j} & \leq |S| - 1 & \forall S \subseteq \custs, \veh \in \vehs}%
      {x \in \mathbb{B}, u \in \mathbb{B}, y \geq 0}

\equationset{Flexible Fleet and Commodity Flow}
\named{fc}%
      {FC}%
      {\sum_{\veh \in \vehs} \sum_{\vtype \in \vtypes} \sum_{\edge_{i, j} \in \edges} x_{\veh, \vtype, i, j} \cdot \mathbf{cost}_\vtype \cdot \mathbf{dist}_{i, j}}%
      {\sum_{\veh \in \vehs} y_{\veh, \cust, \comm} & = \dem_{\cust, \comm} & \forall \cust \in \custs, \comm \in \comms \\
        && \sum_{i \in \custs} \sum_{k \in \comms} y_{v, i, k} & \leq \sum_{t \in \vtypes} z_{v, t} \cdot \capa_{t} & \forall \veh \in \vehs \\
        && \sum_{i \in \custs} y_{v, i, k} & \leq \sum_{t \in \vtypes} z_{v, t} \cdot \capa_t \cdot \comp_{t, k} & \forall \veh \in \vehs, k \in \comms \\
        && \sum_{t \in \vtypes} z_{v, t} &= 1 & \forall \veh \in \vehs \\
        && \sum_{\vtype \in \vtypes} \sum_{\edge_{i,j} \in \edges} x_{\veh, \vtype, i, j} & \leq u_{\veh} & \forall \veh \in \vehs \\
        && \sum_{\edge_{i, j} \in \edges} x_{v, t, i, j} & \leq |\edges| \cdot z_{v, t} & \forall \veh \in \vehs, t \in \vtypes \\
        && y_{v, i, k} & \leq\sum_{t \in \vtypes} \sum_{j \in \custi} x_{v, t, j, i} \cdot  \dem_{i, k} & \forall \veh \in \vehs, i \in \custs, k \in \comms\\
        && \sum_{j \in \custi} (f_{\veh,\comm, j, i} - f_{\veh, \comm, i, j}) & = y_{\veh, i, \comm} & \forall \veh \in \vehs, i \in \custs,  \comm \in \comms \\
        && \sum_{\comm \in \comms} f_{\veh, \comm, i, j} & \leq x_{\veh, \vtype, i, j} \cdot \capa_\vtype & \forall \veh \in \vehs, \edge_{i,j} \in \edges}%
      {x \in \mathbb{B}, u \in \mathbb{B}, y \geq 0, f \geq 0}

\equationset{Flexible Fleet and Vehicle Flow}
\named{ff}%
      {FF}%
      {\sum_{\veh \in \vehs} \sum_{\vtype \in \vtypes} \sum_{\edge_{i, j} \in \edges} x_{\veh, \vtype, i, j} \cdot \mathbf{cost}_\vtype \cdot \mathbf{dist}_{i, j}}%
      {\sum_{\veh \in \vehs} y_{\veh, \cust, \comm} & = \dem_{\cust, \comm} & \forall \cust \in \custs, \comm \in \comms \\
        && \sum_{i \in \custs} \sum_{k \in \comms} y_{v, i, k} & \leq \sum_{t \in \vtypes} z_{v, t} \cdot \capa_{t} & \forall \veh \in \vehs \\
        && \sum_{i \in \custs} y_{v, i, k} & \leq \sum_{t \in \vtypes} z_{v, t} \cdot \capa_t \cdot \comp_{t, k} & \forall \veh \in \vehs, k \in \comms \\
        && \sum_{t \in \vtypes} z_{v, t} &= 1 & \forall \veh \in \vehs \\
        && \sum_{\vtype \in \vtypes} \sum_{\edge_{i,j} \in \edges} x_{\veh, \vtype, i, j} & \leq u_{\veh} & \forall \veh \in \vehs \\
        && \sum_{\edge_{i, j} \in \edges} x_{v, t, i, j} & \leq |\edges| \cdot z_{v, t} & \forall \veh \in \vehs, t \in \vtypes \\
        && y_{v, i, k} & \leq\sum_{t \in \vtypes} \sum_{j \in \custi} x_{v, t, j, i} \cdot  \dem_{i, k} & \forall \veh \in \vehs, i \in \custs, k \in \comms\\
        && \sum_{j \in \custi} x_{\veh, \vtype, i, j} & = \sum_{j \in \custi} x_{\veh, \vtype, j, i} & \forall \veh \in \vehs, \vtype \in \vtypes, i \in \custi \\
        && \sum_{\edge_{i, j} \in \edges^S} x_{\veh, \vtype, i, j} & \leq |S| - 1 & \forall S \subseteq \custs, \veh \in \vehs, \vtype \in \vtypes}%
      {x \in \mathbb{B}, u \in \mathbb{B}, y \geq 0}

\section{Result tables}
\label{ap:results}

In this section, we present intermediate results table highlighting the
progression of the solution quality. First, the results based on the standard
models described in the previous section; then the results with the addition of
valid cuts; and finally the results when using both valid cuts and symmetry
breaking constraints.

The tables are formatted similarly to \Cref{tab:order}: for each instance we
report the best value found; for each model, the solving time, in seconds, or
the gap, in percentage, if the time exceeds $15$ min; for both Vehicle flow
models, the number of sub-tour elimination constraints added.

\ctab{Standard MIP model}
     {standard}
     {*{8}{C}}
     {
     \multirow{2}{*}{Instance} & \multirow{2}{*}{Value} & \multicolumn{4}{c}{Time (s) or gap at 900s} & \multicolumn{2}{c}{\# subtours \eqref{eq:subtours}} \\
     \cmidrule(lr){3-6} \cmidrule(lr){7-8}
     & & sc & sf & fc & ff & sf & ff \\
     \midrule
     \rowcolor{Gray}
     1 & 29458.79 & \bfseries (5.12\%) & (66.38\%) & (6.26\%) & (78.87\%) & 1824 & 2700 \\
     \rowcolor{Gray}
     2 & 40376.14 & \bfseries (5.75\%) & (73.50\%) & (7.59\%) & (82.28\%) & 1764 & 3055 \\
     \rowcolor{Gray}
     3 & 54150.16 & \bfseries (1.86\%) & (77.03\%) & (2.50\%) & (86.83\%) & 5054 & 3825 \\
     \rowcolor{Gray}
     4 & 45435.52 & \bfseries (1.63\%) & (74.47\%) & (3.90\%) & (87.19\%) & 3379 & 4410 \\
     \rowcolor{Gray}
     5 & 39206.94 & \bfseries (3.44\%) & (60.47\%) & (4.98\%) & (80.77\%) & 2652 & 4930 \\
     \rowcolor{Gray}
     6 & 52069.54 & \bfseries (2.56\%) & (76.21\%) & (4.48\%) & (80.76\%) & 6210 & 8500 \\
     \rowcolor{Gray}
     7 & 62864.21 & \bfseries (2.77\%) & (75.58\%) & (4.22\%) & (86.06\%) & 5160 & 4900 \\
     \rowcolor{Gray}
     8 & 23219.88 & \bfseries (8.43\%) & (43.94\%) & (10.03\%) & (75.76\%) & 1044 & 1360 \\
     \rowcolor{Gray}
     9 & 41724.50 & \bfseries (5.89\%) & (71.85\%) & (7.62\%) & (78.96\%) & 2900 & 3010 \\
     \cmidrule{2-6}
     \multicolumn{2}{r}{Avg. gap} & \bfseries 4.16\% & 68.82\% & 5.73\% & 81.94\% \\
     }

\ctab{Summary: valid cuts}
     {cuts}
     {*{8}{C}}
     {
     \multirow{2}{*}{Instance} & \multirow{2}{*}{Value} & \multicolumn{4}{c}{Time (s) or gap at 900s} & \multicolumn{2}{c}{\# subtours \eqref{eq:subtours}} \\
     \cmidrule(lr){3-6} \cmidrule(lr){7-8}
     & & sc & sf & fc & ff & sf & ff \\
     \midrule
     \rowcolor{Gray}
     1 & 31358.95 & \bfseries (2.78\%) & (4.47\%) & (2.85\%) & (4.70\%) & 384 & 350 \\
     \rowcolor{Gray}
     2 & 45538.02 & (2.97\%) & (5.20\%) & \bfseries (1.65\%) & (5.44\%) & 504 & 845 \\
     \rowcolor{Gray}
     3 & 54150.16 & \bfseries (1.17\%) & (6.69\%) & (3.96\%) & (33.38\%) & 1330 & 510 \\
     \rowcolor{Gray}
     4 & 45435.52 & \bfseries (1.22\%) & (2.94\%) & (3.30\%) & (21.11\%) & 744 & 420 \\
     \rowcolor{Gray}
     5 & 54010.05 & \bfseries (3.16\%) & (4.59\%) & (3.85\%) & (6.51\%) & 936 & 425 \\
     \rowcolor{Gray}
     6 & 60964.35 & \bfseries (2.72\%) & (4.93\%) & (4.40\%) & (45.90\%) & 405 & 1300 \\
     \rowcolor{Gray}
     7 & 63045.96 & \bfseries (1.72\%) & (4.62\%) & (4.61\%) & (40.51\%) & 258 & 1900 \\
     8 & 23219.89 & \bfseries 252.93 & 258.06 & (0.87\%) & (4.15\%) & 0 & 160 \\
     \rowcolor{Gray}
     9 & 44738.91 & \bfseries (4.78\%) & (7.42\%) & (7.32\%) & (21.59\%) & 754 & 980 \\
     \cmidrule{2-6}
     \multicolumn{2}{r}{Avg. gap} & \bfseries 2.57\% & 5.11\% & 3.64\% & 20.37\% \\
     }

\ctab{MIP with valid cuts and symmetry breaking constraints}
     {symmetry}
     {*{8}{C}}
     {
     \multirow{2}{*}{Instance} & \multirow{2}{*}{Value} & \multicolumn{4}{c}{Time (s) or gap at 900s} & \multicolumn{2}{c}{\# subtours \eqref{eq:subtours}} \\
     \cmidrule(lr){3-6} \cmidrule(lr){7-8}
     & & sc & sf & fc & ff & sf & ff \\
     \midrule
     1 & 31358.95 & \bfseries 36.75 & 55.17 & 132.64 & 126.41 & 384 & 250 \\
     \rowcolor{Gray}
     2 & 45538.02 & \bfseries (0.38\%) & (2.35\%) & (0.98\%) & (3.10\%) & 532 & 195 \\
     3 & 54150.16 & \bfseries 284.99 & (1.77\%) & (0.32\%) & (3.58\%) & 1140 & 425 \\
     4 & 45435.52 & (0.11\%) & \bfseries 524.05 & (0.77\%) & (1.24\%) & 372 & 630 \\
     \rowcolor{Gray}
     5 & 53989.22 & (1.16\%) & \bfseries (0.15\%) & (1.39\%) & (2.17\%) & 1053 & 680 \\
     \rowcolor{Gray}
     6 & 60964.26 & \bfseries (0.25\%) & (2.88\%) & (1.39\%) & (5.04\%) & 810 & 600 \\
     \rowcolor{Gray}
     7 & 62865.43 & \bfseries (1.45\%) & (2.09\%) & (1.54\%) & (4.29\%) & 559 & 500 \\
     8 & 23219.89 & \bfseries 4.57 & 6.75 & 28.95 & 40.94 & 18 & 80 \\
     \rowcolor{Gray}
     9 & 44738.91 & \bfseries (0.18\%) & (0.99\%) & (0.83\%) & (0.85\%) & 464 & 210 \\
     \cmidrule{2-6}
     \multicolumn{2}{r}{Avg. gap} & \bfseries 0.59\% & 1.70\% & 1.03\% & 2.89\% \\
     }

\end{document}